\documentclass[]{TEAI}
\usepackage{helvet}

\usepackage{amsmath}
\usepackage{natbib}
\usepackage{graphicx}
\usepackage{subcaption}

\usepackage[toc,page,header]{appendix}
\usepackage[utf8]{inputenc}
\usepackage[T1]{fontenc}
\usepackage{hyperref}
\usepackage{url}
\usepackage{booktabs}
\usepackage{amsfonts}
\usepackage{nicefrac}
\usepackage{microtype}
\usepackage{wrapfig}
\usepackage{float}
\usepackage{multirow}
\usepackage{makecell}
\usepackage{siunitx}
\usepackage{colortbl}
\usepackage{algorithm}
\usepackage{algpseudocode}
\usepackage{array}
\usepackage{diagbox}
\usepackage{caption}
\usepackage{pifont}
\usepackage{enumitem}
\usepackage{tabularx}
\usepackage{xcolor}
\usepackage[dvipsnames]{xcolor}

\usepackage{xspace}

\definecolor{deepgreen}{RGB}{0,150,0}
\definecolor{deepyellow}{rgb}{0.8, 0.6, 0.0}

\newcommand{\system}{CalibAll\xspace}
\newcommand{\tick}{\textcolor{deepgreen}{\ding{51}}}
\newcommand{\cross}{\textcolor{red}{\ding{55}}}
\newcommand{\ie}{i.e.\xspace}

\definecolor{cellgreen}{RGB}{220,245,230}
\definecolor{cellyellow}{RGB}{255,243,205}
\definecolor{cellred}{RGB}{250,220,220}

\newcommand{\cellg}[1]{\cellcolor{cellgreen}{#1}}
\newcommand{\celly}[1]{\cellcolor{cellyellow}{#1}}
\newcommand{\cellr}[1]{\cellcolor{cellred}{#1}}

\title{Unify Robot Actions in Camera Frame}

\author{
    Sicheng Xie\textsuperscript{1,2,3},
    Lingchen Meng\textsuperscript{3},
    Zijie Diao\textsuperscript{1},
    Haidong Cao\textsuperscript{1},
    Zhiying Du\textsuperscript{1}, \\
    Shuyuan Tu\textsuperscript{1},
    Jiaqi Leng\textsuperscript{1},
    Qiuyue Wang\textsuperscript{3},
    Mingsheng Li\textsuperscript{3},
    Shuai Bai\textsuperscript{3}, \\
    Zuxuan Wu\textsuperscript{1,2,$\dagger$},
    Yu-Gang Jiang\textsuperscript{1,$\dagger$}
}

\affiliation{
$^1$\mbox{Institute of Trustworthy Embodied AI, Fudan University}\\
$^2$\mbox{Shanghai Innovation Institute}
\quad
$^3$\mbox{Qwen Team, Alibaba Inc.}
}

\abstract{
\begin{abstract}

Cross-embodiment robot learning requires a unified action representation with consistent semantics across robot platforms. Existing representations suffer from platform-specific inconsistencies, while current solutions either maintain embodiment-specific action heads or learn latent action spaces, without fundamentally resolving the mismatch. We propose to unify robot actions in the camera frame using camera extrinsics, so that actions share consistent geometric semantics across different robot embodiments, including both single-arm and bimanual robots. However, most existing datasets lack camera extrinsic annotations, and existing offline calibration methods either suffer from local minima or require robot-specific training data. To address this gap, we present \system, a training-free, robot-independent annotation pipeline that estimates camera extrinsics for offline datasets and converts heterogeneous robot actions into standardized camera-frame actions. \system follows a coarse-to-fine calibration strategy: temporal PnP provides a stable initialization, followed by differentiable rendering-based refinement for high precision. Beyond extrinsics, \system produces standardized TCP-pose actions and auxiliary annotations. We apply \system to 16 datasets across 4 robot platforms, producing approximately 97K calibrated data episodes. Downstream simulation and real-robot experiments show that cross-embodiment pretraining with camera-frame actions achieves state-of-the-art performance.
\end{abstract}
}

\checkdata[Code]{\url{https://github.com/SII-dannyXSC/CalibAll}}

\begin{document}
\maketitle
\renewcommand{\thefootnote}{}
\footnotetext{$^\dagger$Corresponding author.}
\renewcommand{\thefootnote}{\arabic{footnote}}

\vspace{-1.5em}

\section{Introduction}
\label{sec:intro}
\begin{figure}[th]
\centering
\includegraphics[width=\linewidth]{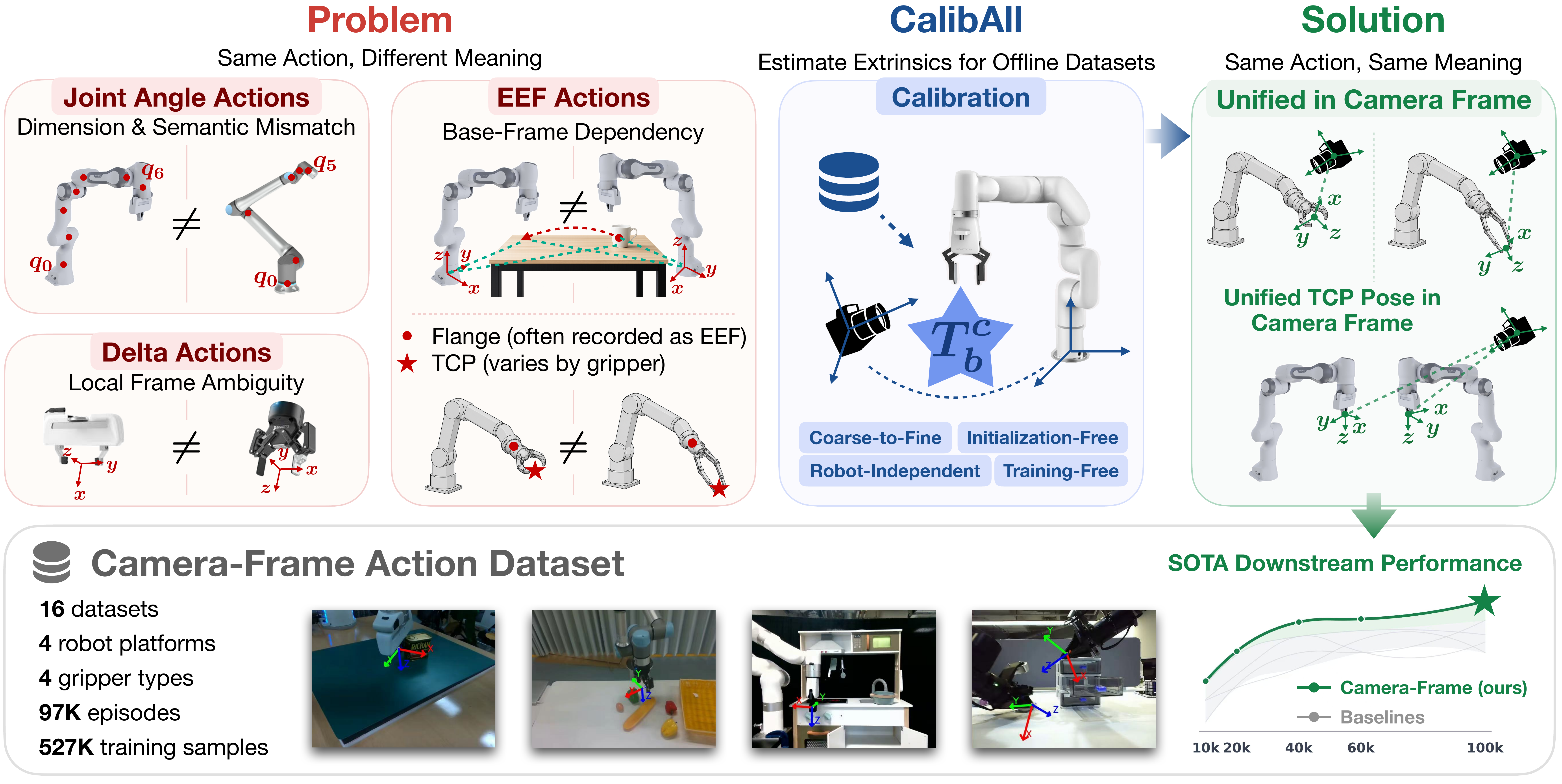}
\caption{Current action representations suffer from inconsistent semantics across robot datasets: joint actions vary in dimensionality, end-effector poses depend on robot-specific base frames and EEF definitions, and delta actions use inconsistent local frames. \system resolves these inconsistencies by estimating camera extrinsics for offline datasets and converting heterogeneous actions into standardized TCP poses in the camera frame. We use this pipeline to build a Camera-Frame Action Dataset for cross-embodiment pretraining, leading to state-of-the-art downstream performance.}
\label{fig:teasers}
\end{figure}
Enabling robots to learn and generalize across different embodiments is a central challenge in scaling robot learning~\cite{wang2024scaling, zha2026lap, black2024pi_0}. A core requirement for cross-embodiment learning is a unified action representation~\cite{zheng2025x, kim2024openvla, black2024pi_0} that carries consistent semantics regardless of the robot platform. However, this is far from trivial, as different robots have vastly different kinematic structures, joint configurations, and coordinate conventions.

Existing action representations each have significant limitations for cross-embodiment unification: joint angles are incompatible in dimensionality and semantics across robots, end-effector poses depend on the robot base frame and are often recorded at the flange rather than the more grasp-relevant tool center point (TCP) pose, and delta actions suffer from inconsistent local coordinate conventions.
To circumvent these issues, prior work either employs embodiment-specific action heads~\cite{bjorck2025gr00t, team2024octo, wang2024scaling} that maintain separate output spaces per robot, or learns shared latent actions~\cite{ye2024latent, bu2025univla, zheng2025universal} that abstract away low-level differences. However, these approaches either sidestep the unification problem or require embodiment-specific adaptation, rather than resolving the inconsistency fundamentally.

To address this problem, we propose to unify robot actions into a semantically and formally consistent space by transforming them into the \textbf{camera frame}. Given the camera extrinsic, \ie, the rigid transformation between the robot base and the camera, actions can be expressed in a visual coordinate system shared by the observation. This formulation naturally handles both single-arm and bimanual robots: for single-arm systems, the camera frame provides a consistent reference that is independent of base placement; for bimanual systems, since both arms are observed by the same camera, the camera frame implicitly encodes their spatial relationship. Since the camera frame is defined by the visual observation itself, actions expressed in this space carry consistent geometric meaning across robot configurations, naturally resolving the inconsistencies above.

However, most existing robot manipulation datasets~\cite{o2024open, liu2024rdt, wu2024robomind} lack camera extrinsic annotations, preventing camera-frame action representations from being applied at scale. Existing offline calibration approaches attempt to recover camera extrinsics from data, either by using differentiable rendering to align rendered robot masks with observed masks~\cite{chen2023easyhec,hong2024easyhec++}, or by detecting robot-specific keypoints and solving via Perspective-n-Point (PnP)~\cite{lu2023markerless, lu2025ctrnet}. The former is prone to local minima with poor initialization. The latter requires dense, robot-specific annotations, preventing it from scaling to diverse platforms.

To overcome these limitations, we present \system, a training-free, robot-independent pipeline that estimates camera extrinsics for offline datasets and converts robot actions from heterogeneous sources into standardized camera-frame actions. \system follows a coarse-to-fine pipeline to calibration: in the coarse stage, we track the marked point on the end-effector across the video and solve for extrinsics via temporal PnP. In the fine stage, we refine the estimate through differentiable rendering-based optimization that aligns rendered and observed robot masks, achieving high precision. After calibration, \system computes the tool center point (TCP) pose using the robot arm and gripper URDF and standardizes the local action frame across different grippers, yielding a unified grasp-relevant TCP pose representation in the camera frame.

Using \system, we annotate camera extrinsics for 16 datasets spanning 4 robot platforms including both single-arm and bimanual robots (Franka, UR5e, XArm7, AgileX ALOHA), producing approximately 97K calibrated data episodes. Downstream experiments in both simulation and the real world show that cross-embodiment pretraining with camera-space actions achieves state-of-the-art performance compared with baselines.

Our contributions are threefold:
\begin{itemize}
    \setlength{\leftskip}{-2em}
    \setlength{\itemsep}{0.2em}
    \setlength{\topsep}{-0.2em}
    
    \item We propose camera-frame actions as a unified action representation for cross-embodiment robot learning. By expressing actions in the visual coordinate system and using a standardized TCP-pose representation, our formulation provides consistent geometric semantics across different robot configurations, including both single-arm and bimanual robots.

    \item We introduce \system, a training-free and robot-independent annotation pipeline that estimates camera extrinsics for offline datasets and converts heterogeneous robot actions into standardized camera-frame actions. \system uses a coarse-to-fine calibration strategy and further produces grasp-relevant TCP poses and standardized local action frames.

    \item We apply \system to 16 robot manipulation datasets across 4 robot platforms, producing approximately 97K calibrated episodes, together with auxiliary annotations including robot masks, bounding boxes, and 2D/3D trajectories. Downstream simulation and real-robot experiments show that cross-embodiment pretraining with camera-frame actions outperforms baselines.
\end{itemize}

\section{Related Work}

\paragraph{Extrinsic Calibration.}
Hand-eye calibration estimates the relative pose between the camera and the robot. In this work, we focus on the eye-to-hand setting, where an external camera observes the robot and the goal is to estimate the camera extrinsic with respect to the robot base. Traditional online methods rely on markers or require access to the physical robot during calibration~\cite{ilonen2011robust,tsai1989new,park1994robot,daniilidis1999hand}. Recent markerless methods such as EasyHeC~\cite{chen2023easyhec} and EasyHeC++~\cite{hong2024easyhec++} use differentiable rendering to align rendered robot masks with observed images, but they still require robot access and are sensitive to initialization. Offline methods such as DREAM~\cite{lee2020camera}, CtRNet~\cite{lu2023markerless}, and CtRNet-X~\cite{lu2025ctrnet} estimate camera poses from collected images, but rely on robot-specific keypoint annotations or training data. In contrast, \system targets offline datasets and estimates camera extrinsics in a training-free and robot-independent manner through a coarse-to-fine pipeline.

\paragraph{Cross-Embodiment Action Representation.}
Large robot policies are often trained on heterogeneous datasets collected from different embodiments, making action unification critical. Joint-angle actions are difficult to share across robots because different platforms have different numbers of joints and platform-specific joint semantics~\cite{fu2024mobile,zhao2023learning}. End-effector actions provide a unified dimensionality, but they are usually defined in robot-specific base frames and are often recorded at the flange rather than the grasp-relevant tool center point (TCP)~\cite{intelligence2025pi_,black2024pi_0,o2024open,zitkovich2023rt,kim2024openvla}. Delta actions reduce dependence on global coordinates, but their local frame conventions can still vary across robots. Existing approaches either use embodiment-specific action heads~\cite{bjorck2025gr00t,team2024octo,wang2024scaling} or learn shared latent actions~\cite{ye2024latent,bu2025univla,zheng2025universal}, which alleviate but do not fundamentally resolve the inconsistency of low-level action semantics.

\paragraph{Camera-Space Actions.}
Camera-space actions transform robot actions into the camera frame using camera extrinsics, providing a unified geometric reference shared with visual observations. Recent works have used camera-space actions for human-to-robot learning, bimanual manipulation, and viewpoint-robust VLA policies~\cite{wang2023mimicplay,lepert2025phantom,fang2025airexo,jiang2025you,zhang2025grounding}. However, most large-scale robot datasets do not provide accurate camera extrinsics, which prevents camera-space actions from being applied broadly. Our work addresses this bottleneck by introducing a scalable annotation pipeline that estimates camera extrinsics for offline datasets and converts heterogeneous robot actions into standardized camera-frame TCP actions.

\section{Method}
\label{sec:method}
We present the \system annotation pipeline for unifying robot actions into the camera space for offline manipulation datasets. The pipeline consists of three stages: preprocessing (\cref{sec:preprocess}), camera extrinsic estimation (\cref{sec:extrinsic}), and postprocessing (\cref{sec:postprocess}). We then apply this pipeline to construct a large-scale Camera-Frame Action Dataset (\cref{sec:dataset}).

\subsection{Problem Setup}
\label{sec:task_formulation}

Given an offline robot manipulation dataset, the goal of \system is to convert all robot actions into a unified camera-frame representation with consistent semantics across different embodiments. As discussed in~\cref{sec:intro}, this requires the camera extrinsic $T_b^c$, i.e., the rigid transformation from the robot base frame $b$ to the camera frame $c$, satisfying $P_c = T_b^c P_b$. Here, $P_b$ denotes a 3D point in the robot base frame, and $P_c$ denotes the same physical point in the camera frame. Estimating $T_b^c$ accurately and at scale is therefore the central technical challenge of our pipeline.

The pipeline requires only two types of information that are commonly available in existing datasets: (1) RGB video sequences, (2) Joint angles at each timestamp.

\subsection{Preprocessing}
\label{sec:preprocess}
Before running the calibration pipeline, several preprocessing steps are required.
\paragraph{Camera Intrinsic Estimation.}
Camera intrinsics $K$ are needed for camera extrinsic estimation (\cref{sec:extrinsic}). When the dataset provides intrinsics directly, we use them as-is. Otherwise, we look up the intrinsics from the known camera model (e.g., RealSense D435, ZED 2). If neither is available, we estimate the intrinsics using off-the-shelf methods~\cite{wang2025moge}.
\paragraph{Video Clip Selection.}
The coarse initialization stage (\cref{sec:coarse_init}) relies on tracking the end-effector throughout the video. If the gripper leaves the camera field of view during the sequence, the tracker may lose the target and produce erroneous trajectories. To ensure tracking quality, we manually select a video clip for each scene in which the gripper remains fully visible. This clip is used for calibration, while the resulting extrinsic applies to the entire scene since the camera-robot relative pose is fixed.
\paragraph{Bimanual Robot Handling.}
For bimanual robots where both arms are defined in a single URDF with a fixed spatial relationship, we only need to calibrate one arm; the extrinsic of the other arm is automatically determined by the known inter-arm transform. For setups where the two arms are independently mounted and their relative position is not known, we calibrate each arm separately by estimating an independent extrinsic $T_{b_i}^c$ for each arm base $b_i$. In both cases, since both arms are observed by the same camera, the camera frame naturally encodes their spatial relationship.

\subsection{Camera Extrinsic Estimation}
\label{sec:extrinsic}
\begin{figure}[t]
\centering
\includegraphics[width=\textwidth]{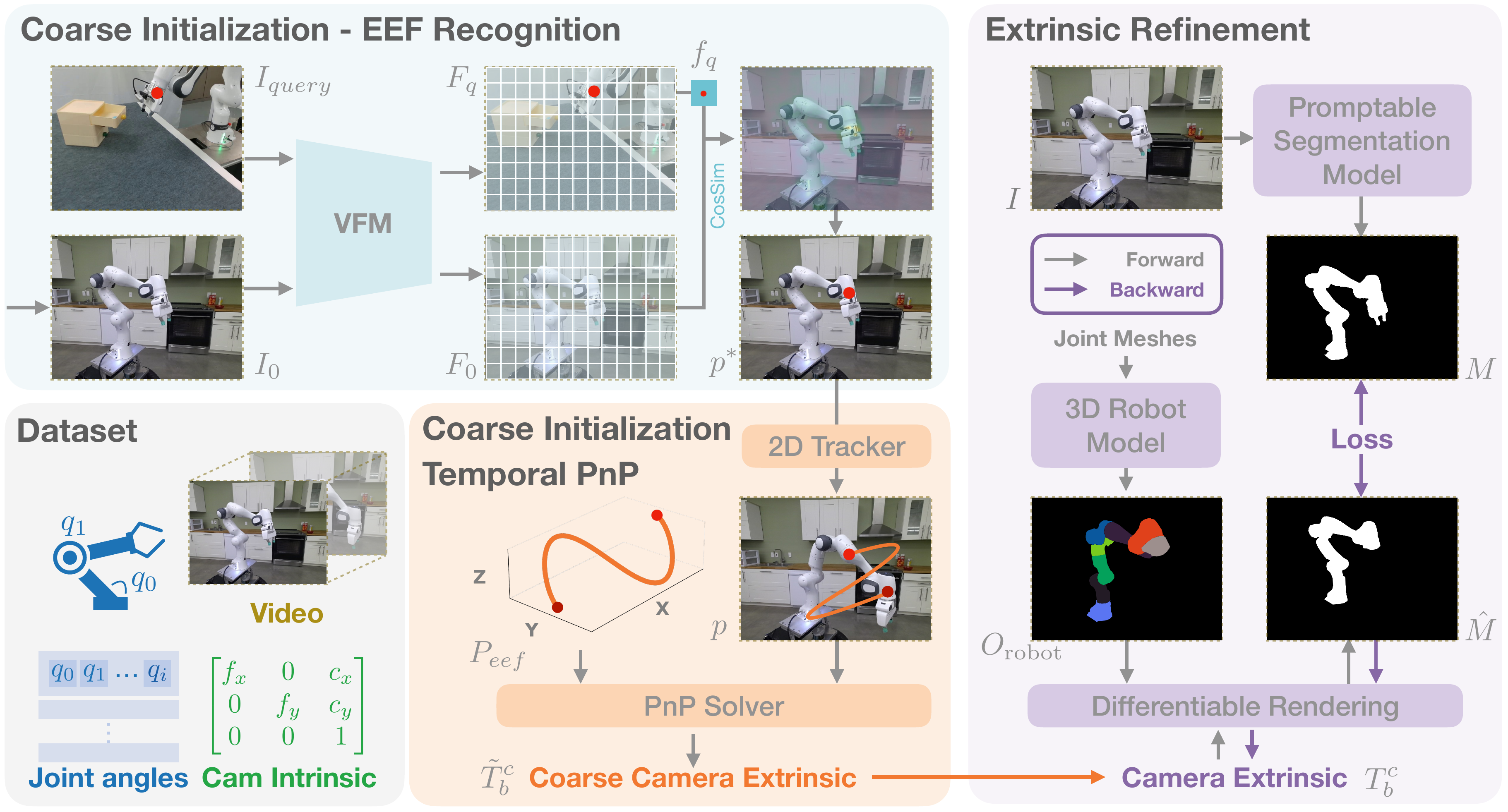}
\caption{\textbf{Overview of \system.}
It follows a coarse-to-fine calibration pipeline that achieves training-free, stable, and accurate camera extrinsic estimation across diverse datasets and robot platforms. \system first use EEF Recognition to obtain the end-effector tracking point, then apply temporal PnP to estimate a coarse extrinsic, and finally perform extrinsic refinement to obtain an accurate result.
}
\label{fig:framework}
\vspace{-0.5em}
\end{figure}

We estimate the camera extrinsic $T_b^c$ through a coarse-to-fine approach: a temporal PnP stage produces a stable initial estimate, which is then refined via differentiable rendering-based optimization.

\subsubsection{Coarse Initialization}
\label{sec:coarse_init}

\paragraph{EEF Recognition.}
The coarse initialization stage requires a tracking point $p^*$ on the end-effector (EEF). Although this point can be manually annotated, we explore an automatic alternative using cross-image correspondence from a vision foundation model (VFM)~\cite{tang2023emergent,meng2024segic}. Given a reference mark $p_q$ on an EEF image, we extract its feature and find the most similar location in the first frame $I_0$:
\begin{equation}
p^* = \arg\max_{(u,v)} \ \mathrm{Cos}\Big({f}_{q}, {F}_0(u,v)\Big),
\end{equation}
where ${f}_{q}$ and ${F}_0$ are VFM features of the reference mark and the target frame, respectively. Since EEFs share common visual characteristics across robots, this enables stable localization across diverse platforms with a single mark. Representative localization examples are shown in~\cref{fig:point_result_vis}.

\paragraph{Temporal PnP.}
Using a 2D point tracking model~\cite{karaev2024cotracker,karaev2025cotracker3} with the mark $p^*$ as input, we track the EEF throughout the video to obtain a 2D trajectory $p$. Meanwhile, we compute the corresponding 3D EEF trajectory $P_{{eef}}$ via forward kinematics from the joint angles $Q$ and the robot model $O$. An PnP solver~\cite{lepetit2009ep} then estimates the camera extrinsic:
\begin{equation}
\tilde{T}_b^{c} = \mathrm{PnP}\left(p,P_{eef},K\right).
\end{equation}
Since $p^*$ may not precisely correspond to the 2D projection of $P_{eef}$ and the tracked trajectory contains inevitable noise, the result is a coarse approximation. Nevertheless, it reliably captures the camera position and viewing direction, serving as an effective initialization for the subsequent refinement. We provide qualitative visualizations in~\cref{fig:coarse_refine} and quantitative ablations in~\cref{tab:ablation}.

\subsubsection{Rendering-based Refinement}
\label{sec:refinement}

We refine the coarse estimate via rendering-based optimization that exploits the full robot geometry. The key idea is to iteratively adjust the camera pose until the rendered robot mask aligns with the observed one, following the render-and-compare paradigm~\cite{kundu20183d}.

We employ a promptable segmentation foundation model~\cite{carion2025sam,liu2024grounding,ren2024grounding,ravi2024sam} to obtain the robot arm mask $M$ for each image. We then adopt a differentiable rendering pipeline~\cite{laine2020modular} to render the corresponding robot mask $\hat{M}$ under the current camera pose:
\begin{equation}
\hat{M} =  \min\!\left(1,\ \sum_{l} \pi\!\left(\boldsymbol{T_{b}^{c}}\, T_{l}^{b}l\right)\right),
\end{equation}
where $\pi$ is the differentiable mask renderer, $l$ denotes a robot link, and $T_l^b$ is the link-to-base transformation computed via forward kinematics. Masks are rendered per link and aggregated, with the min operator ensuring valid binary occupancy.

The camera extrinsic $\boldsymbol{T}_b^{c}$ is optimized via gradient descent:
\begin{equation}
L\!\left(T_{b}^{c}\right)
= \left\| M - \hat{M} \right\|_2 
= \left\| M - \min\!\left(1,\ \sum_{l} \pi\!\left(\boldsymbol{T_{b}^{c}}\, T_{l}^{b}l\right)\right) \right\|_2 .
\end{equation}

Rendering-based optimization is prone to local minima with poor initialization (\cref{tab:mask_result}). If the initial camera pose fails to capture the robot, the rendered mask collapses to zero, producing no gradients and halting optimization. The coarse initialization from~\cref{sec:coarse_init} is therefore essential for stable convergence (\cref{tab:ablation}, \cref{tab:mask_result}).

\subsection{Postprocessing}
\label{sec:postprocess}

After obtaining the camera extrinsic $T_b^c$, we perform several postprocessing steps to produce the final annotations.

\paragraph{Standardized Camera-Frame Action Conversion.}
The primary output of our pipeline is a unified camera-space action representation. This conversion consists of three steps:
(1) Convert the raw flange EEF pose into a tool center point (TCP) pose. Since different grippers have different geometries, the flange pose does not necessarily correspond to the actual grasping point. We define the action position as the TCP, located at the tip of the closed gripper fingers, and obtain the flange-to-TCP offset $T_{\mathrm{tcp}}^{\mathrm{flange}}$ from the gripper URDF.
(2) Standardize the TCP local frame. We introduce a gripper-specific transform $S_{\mathrm{tcp}}$ such that the $z$-axis points along the gripper forward direction and the $y$-axis points along the gripper closing direction. This makes the TCP rotation semantics consistent across robot platforms.
(3) Transform the standardized TCP pose from the robot base frame to the camera frame:
\begin{equation}
T_{\mathrm{tcp}}^{c}
=
T_b^c\,
T_{\mathrm{flange}}^{b}\,
T_{\mathrm{tcp}}^{\mathrm{flange}}\,
S_{\mathrm{tcp}},
\end{equation}
where $T_b^c$ is the camera extrinsic, $T_{\mathrm{flange}}^{b}$ is the flange pose in the robot base frame computed by forward kinematics, $T_{\mathrm{tcp}}^{\mathrm{flange}}$ is the flange-to-TCP transform obtained from the gripper URDF, and $S_{\mathrm{tcp}}$ is the gripper-specific transform that standardizes the TCP local frame. For bimanual robots, we apply the same conversion independently to each arm using its own extrinsic $T_{b_i}^c$, so that both arms are represented in the shared camera frame.

\paragraph{Unified Action Dimension.}
To enable joint training of single-arm and bimanual data, we unify the action dimension across all robots. Each arm produces a 7-dimensional camera-frame action (3D position + 3D rotation + 1D gripper). For single-arm robots, we pad the action to 14 dimensions by appending zeros for the second arm. For bimanual robots, the actions of both arms are concatenated into the same 14-dimensional vector. This simple padding scheme allows single-arm and bimanual data to be mixed in a single training batch without architectural changes.

\paragraph{Auxiliary Annotations.}
Beyond camera-frame actions, \system also produces auxiliary annotations, including masks and bounding boxes for both the robot and gripper, 3D TCP trajectories in the camera frame, and corresponding 2D TCP trajectories in the image space as shown in~\cref{sec:vis_dataset}. These annotations are generated using the estimated camera extrinsics and robot URDF. Specifically, the URDF provides the robot geometry and kinematic structure, while $T_b^c$ aligns the robot model with the camera frame. We render the robot in the image space to obtain masks and bounding boxes, and project the 3D TCP trajectory onto the image plane to obtain 2D trajectories. These annotations can support downstream tasks such as visual grounding~\cite{liu2024grounding}, robot-aware policy learning~\cite{feng2025vidar}, and trajectory-conditioned manipulation~\cite{tan2026robobrain}.

\subsection{Large-scale Dataset Construction}
\label{sec:dataset}
We apply \system to a large collection of existing robot manipulation datasets to construct the Camera-Frame Action Dataset.
Although \system reduces the need for robot-specific calibration and training data, fully automatic annotation remains challenging for large-scale offline datasets in practice. Visual tracking may fail under severe self-occlusion, and foundation-model masks can be inaccurate in cluttered scenes or under challenging viewpoints. To ensure annotation quality, we adopt a human-in-the-loop protocol: when tracking fails, we manually provide suitable initial tracking points instead of relying solely on automatic EEF localization; when mask prediction is inaccurate, we manually select the correct mask. We leave fully automatic large-scale offline calibration under arbitrary real-world conditions as future work.

As shown in~\cref{tab:dataset_summary}, the resulting dataset is built from three major data sources: RoboMIND~\cite{wu2024robomind}, RDT~\cite{liu2024rdt}, and Open X-Embodiment (OXE)~\cite{o2024open}. It covers 16 datasets, 4 robot platforms, and 4 gripper types, including both single-arm robots (Franka, UR5e, and XArm7) and bimanual robots (AgileX ALOHA). In total, the dataset contains approximately 97K episodes and 527K frame-level training samples. Since camera extrinsics are annotated per third-person camera view, an episode with multiple third-person cameras contributes multiple calibrated samples. Each calibrated sample includes the camera extrinsic $T_b^c$, standardized camera-frame TCP actions, and the auxiliary annotations described in~\cref{sec:postprocess}.

\section{Experiments}
\label{sec:experiments}
\subsection{Calibration Experiments}

We first evaluate the calibration accuracy of \system by comparing with existing offline methods.

\paragraph{Experimental Setup.}
We evaluate camera extrinsic estimation on the DREAM dataset~\cite{lee2020camera}, which contains approximately 57K real-world images of the Franka Panda robot. We compare \system with existing offline calibration methods, including DREAM~\cite{lee2020camera} variants (F, Q, H), RoboPose~\cite{labbe2021single}, CtRNet~\cite{lu2023markerless}, and CtRNet-X~\cite{lu2025ctrnet}. We report Average Distance (ADD) and Area-Under-the-Curve (AUC). Given 3D points $\{p_i\}_{i=1}^{n}$ in the robot base frame, we compute their alignment error under the predicted extrinsic $\tilde{T}_b^c$ and ground-truth extrinsic $T_b^c$:
\begin{equation}
D_i=\left\| \tilde{T}_b^c p_i - T_b^c p_i \right\|_2, 
\quad
\mathrm{ADD} = \frac{1}{n}\sum_{i=1}^n D_i,
\quad
\mathrm{AUC} = \frac{1}{nT}\sum_{t=1}^{T}\sum_{i=1}^{n} \mathbf{1}(D_i < \tau_t),
\label{equ:add_auc}
\end{equation}
where $\tau_t$ is the $t$-th distance threshold and $T$ is the number of thresholds.

\paragraph{Results.}
\begin{table}[t]
\caption{Calibration Results on DREAM~\cite{lee2020camera} and comparison with offline calibration methods.}
\label{tab:main}
\resizebox{\linewidth}{!}{
    \begin{tabular}{l|cc|cccc}
    \toprule
             & AUC $\uparrow $ & ADD (m) $\downarrow$ & Train-Free  & Init-Free & Robot in-and-out & Robot-Independent \\
    \midrule
    \textit{DREAM-F~\cite{lee2020camera}}    & 60.740       & 113.029      & \cross & \tick  & \cross    & \cross    \\
    \textit{DREAM-Q~\cite{lee2020camera}}    & 56.988       & 59.284       & \cross & \tick  & \cross    & \cross    \\
    \textit{DREAM-H~\cite{lee2020camera}}    & 68.584       & 17.477       & \cross & \tick  & \cross    & \cross    \\
    \textit{RoboPose~\cite{labbe2021single}} & 80.094       & 0.020        & \cross & \cross & \cross    & \cross    \\
    \textit{CtRNet~\cite{lu2023markerless}}  & 85.962       & 0.020        & \cross & \tick  & \cross    & \cross    \\
    \textit{CtRNet-X~\cite{lu2025ctrnet}}    & 86.231       & 0.014        & \cross & \tick  & \tick     & \cross    \\
    \textit{\textbf{\system(ours)}}          & \textbf{97.642} & \textbf{0.008}     & \tick  & \tick  & \tick     & \tick     \\
    \bottomrule
    \end{tabular}}
\end{table}

As shown in~\cref{tab:main}, \system achieves an AUC of 97.642 and ADD of 0.008 m, substantially outperforming all baselines. The best prior method, CtRNet-X, reaches 86.231 AUC and 0.014 m ADD. \system improves the AUC by over 11 points while being the only method that is simultaneously training-free, and robot-independent.

\begin{wraptable}{r}{0.48\linewidth}
    \centering
    \small
    \vspace{-1em}
    \setlength{\tabcolsep}{3pt}
    \caption{Ablation study on DREAM~\cite{lee2020camera}.}
    \label{tab:ablation}
    \begin{tabular}{l|cc}
        \toprule
        & AUC $\uparrow$ & ADD (m) $\downarrow$ \\
        \midrule
        Coarse only & 82.050 & 0.071 \\
        Fine only & 0 & 0.5+ \\
        \rowcolor{gray!20}
        Coarse + Fine & \textbf{97.642} & \textbf{0.008} \\
        \bottomrule
    \end{tabular}
\end{wraptable}
\paragraph{Ablation study.}
We ablate the two stages of the camera extrinsic estimation pipeline on the DREAM dataset (\cref{tab:ablation}). Using coarse initialization alone achieves an AUC of 82.050 and ADD of 0.071 m, indicating that temporal PnP already provides a reasonably good estimate. Using rendering-based refinement alone with random initialization fails completely (AUC of 0), confirming that the optimization is prone to local minima without a good starting point. The full pipeline (coarse + refine) achieves the best result, demonstrating that both stages are essential and complementary.

\begin{wraptable}{r}{0.58\linewidth}
\centering
\small
\vspace{-1em}
\caption{Convergence regions of extrinsic refinement.}
\label{tab:mask_result}
\vspace{-0.5em}
\setlength{\tabcolsep}{3pt}
\renewcommand{\arraystretch}{1.15}
\begin{tabular}{c|ccccc}
\toprule
\multirow{2}{*}{\textbf{Rot (°)}} &
\multicolumn{5}{c}{\textbf{Pos Noise (cm)}} \\
\cmidrule(lr){2-6}
 & 0.1--2.5 & 2.5--5 & 5--7.5 & 7.5--10 & 10--15 \\
\midrule
1--5   & \cellg{60/60} & \cellg{60/60} & \cellg{56/60} & \celly{41/60} & \cellr{19/60} \\
5--10  & \cellg{60/60} & \cellg{58/60} & \cellg{49/60} & \cellr{25/60} & \cellr{16/60} \\
10--15 & \cellg{54/60} & \cellg{47/60} & \celly{38/60} & \cellr{24/60} & \cellr{16/60} \\
15--20 & \celly{39/60} & \celly{36/60} & \cellr{16/60} & \cellr{18/60} & \cellr{11/60} \\
20--25 & \cellr{28/60} & \cellr{21/60} & \cellr{14/60} & \cellr{19/60} & \cellr{6/60} \\
\bottomrule
\end{tabular}
\vspace{-1em}
\end{wraptable}
\paragraph{Convergence analysis of refinement.}
To further investigate the convergence behavior of the refinement stage, we conduct a controlled experiment (\cref{tab:mask_result}). We select ground-truth camera extrinsics, add varying levels of rotational and positional noise, and run the refinement from these noisy initializations. For each noise level, we conduct 60 trials and record the number of successful convergences. As shown in~\cref{tab:mask_result}, when the initialization error is within approximately 10° rotation and 7.5 cm translation, the refinement converges reliably (green region). Beyond this range, the success rate drops sharply (red region). Notably, our coarse initialization stage produces estimates with an average rotation error of 4.0° and position error of 5.5 cm on the DREAM dataset, both well within the convergence region, explaining why the full pipeline achieves stable results.

\paragraph{Visualization results.}
We test \system on DROID~\cite{khazatsky2024droid} in~\cref{sec:droid_result}, a dataset with provided camera extrinsics, and find that \system provides more precise calibration results than the original dataset annotations, yielding better alignment between the rendered robot model and the observed image.
We also visualize the EEF recognition results in~\cref{fig:point_result_vis}, as well as the rendered robot models obtained using the camera extrinsics estimated by the coarse and refinement stages as shown in~\cref{fig:coarse_refine}. 
We additionally visualize 10 representative cases from our Camera-Frame Action Dataset as shown in~\cref{sec:vis_dataset}, including projected 2D TCP trajectories, standardized local frames, and rendered bounding boxes and masks. These examples demonstrate that \system produces accurate and scalable annotations across diverse robot platforms, camera viewpoints, and manipulation scenes.

\subsection{Simulation Experiments}
\label{sec:sim_exp}
\paragraph{Pretraining Variants.}
We compare several pretraining variants to study the effect of unifying action representations across embodiments:
\begin{itemize}
    \setlength{\leftskip}{-2em}
    \setlength{\itemsep}{0.2em}
    \setlength{\topsep}{-0.2em}

    \item \textbf{From Scratch}: no cross-embodiment pretraining; the model is trained only with SFT on the target benchmark.
    \item \textbf{Raw Mix}: cross-embodiment pretraining by directly mixing datasets with their original action representations. Actions from different datasets are padded to the same dimensionality.
    \item \textbf{Prompt}: cross-embodiment pretraining by specifying robot embodiment information in the language prompt, while padding the original action representations to the same dimensionality without semantic unification as shown in~\cref{sec:training_details}.
    \item \textbf{Action Head}: cross-embodiment pretraining with dataset-specific action heads, where each dataset uses a separate head to encode and decode its own action representation.
    \item \textbf{Camera-Frame (ours)}: cross-embodiment pretraining using camera-frame actions derived from \system annotations, yielding a unified action representation across different robot platforms.
\end{itemize}
\begin{table}[t]
\centering
\caption{Success rate (\%) on RobotTwin~\cite{chen2025robotwin} Clean.}
\label{tab:downstream}
\begin{tabular}{l|cccccc|c}
\toprule
\textbf{Pretrain Type} & \textbf{10k} & \textbf{20k} & \textbf{40k} & \textbf{60k} & \textbf{80k} & \textbf{100k} & \textbf{Best} \\
\midrule
From Scratch    & 12.8 & 23.5 & 32.9 & 33.0 & 31.2 & 42.4 & 42.4 \\
Raw Mixed  & 19.4 & 26.6 & 37.7 & 33.3 & \textbf{33.6} & 36.1 & 37.7 \\
Prompt  & 23.7& 32.3& 37.7& 39.4& 29.4& 38.2& 39.4\\
Action Head  & 19.1 & 22.8 & 28.5& 36.8& 31.4& 33.0& 36.8 \\
\rowcolor{gray!15}
Camera-Frame (ours) & \textbf{24.7} & \textbf{33.4} & \textbf{41.9} & \textbf{42.8} & 29.7 & \textbf{48.0} & \textbf{48.0} \\
\bottomrule
\end{tabular}

\end{table}

\paragraph{Experimental Setup.}
We use StarVLA~\cite{community2026starvla} as the training framework. All variants are evaluated on the RoboTwin benchmark. We use the RoboTwin~\cite{chen2025robotwin} as benchmark, which contains 2,500 trajectories for training. All baselines are pretrained on the original datasets, whereas \textbf{Camera-Frame} is pretrained on our Camera-Frame Action Dataset. 

\paragraph{Camera-frame pretraining achieves the best performance.}
As shown in~\cref{tab:downstream}, \textbf{Camera-Frame} achieves the highest success rate in 5 out of 6 SFT checkpoints, reaching the best overall performance of 48.0\% at 100k steps. This improves over From Scratch by 5.6\%, Raw Mix by 10.3\%, Prompt by 8.6\%, and Action Head by 11.2\% in terms of best success rate.

\paragraph{Naive cross-embodiment pretraining can lead to negative transfer.}
Raw Mix performs worse than From Scratch at convergence, with 36.1\% versus 42.4\% at 100k steps. This suggests that directly mixing data from different robots without action representation unification can introduce negative transfer, since the model must fit inconsistent action semantics under a shared action space. Action Head also underperforms From Scratch in terms of best success rate, achieving only 36.8\%, indicating that using dataset-specific heads alone does not fully resolve the underlying inconsistency. Prompt improves over Raw Mix and Action Head, reaching 39.4\% best success rate, but still falls behind From Scratch and Camera-Frame, suggesting that embodiment information in the language prompt is insufficient to eliminate action-space misalignment.

\paragraph{Camera-frame pretraining maintains strong performance across SFT steps.}
Camera-Frame consistently performs well throughout the SFT process, achieving the best results at 10k, 20k, 40k, 60k, and 100k steps. This stable advantage suggests that the unified camera-frame action representation provides a more transferable pretraining signal across embodiments. In particular, Camera-Frame already reaches 24.7\% at 10k steps and continues to improve to 48.0\% at 100k steps, indicating both faster adaptation and stronger final performance.

\subsection{Real-Robot Experiments}
\begin{figure}[th]
\centering
\includegraphics[width=\linewidth]{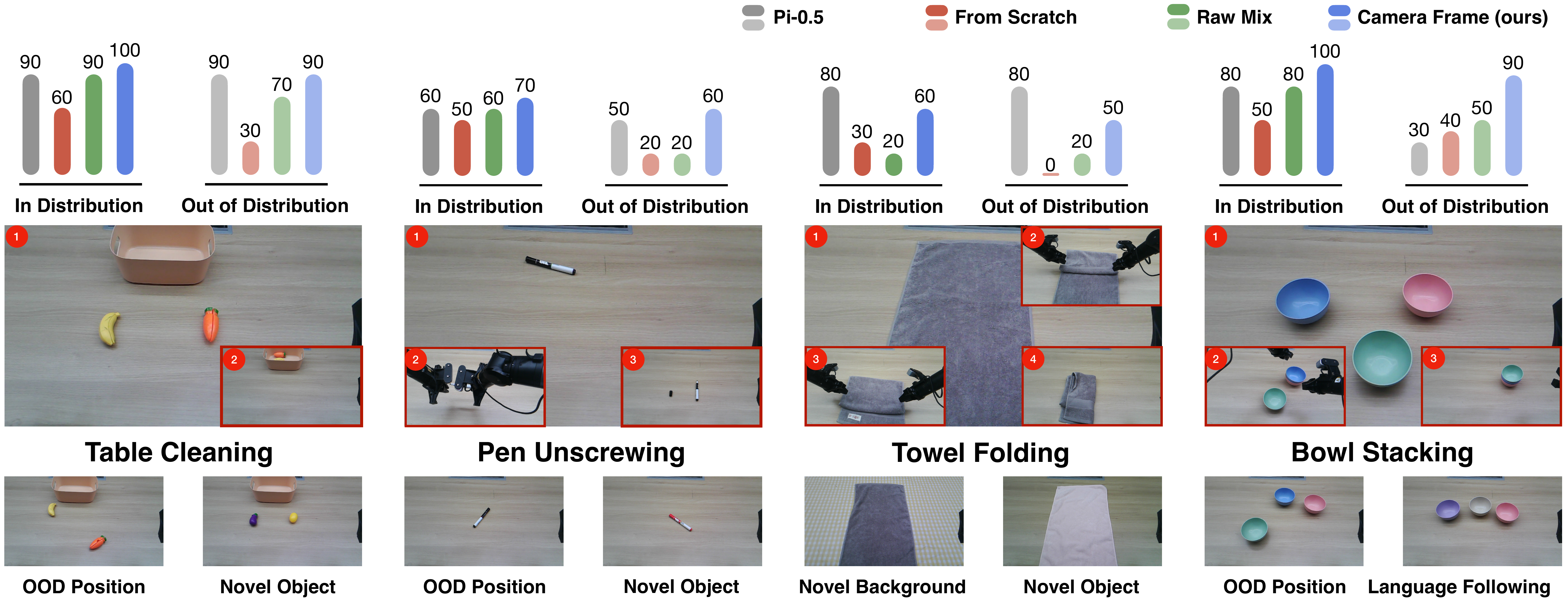}
\caption{\textbf{Real-World Experiments.} The top charts report results from 10 in-distribution (ID) trials and 10 out-of-distribution (OOD) trials. The bottom panel illustrates the OOD test setup.}
\label{fig:real_exp}
\end{figure}

\paragraph{Experimental Setup.}
We evaluate real-world closed-loop performance in a few-shot tabletop setting on an ALOHA platform. The policy observes one third-person camera, two wrist cameras, the task prompt, and proprioceptive states. We compare four variants: From Scratch, Raw Mix, Camera-Frame, and $\pi_{0.5}$~\cite{intelligence2025vision} as a reference pretrained policy.

We evaluate four tasks: Table Cleaning, Pen Unscrewing, Towel Folding, and Bowl Stacking as shown in~\cref{fig:real_exp}. Each task includes both in-distribution and OOD settings. The OOD tests introduce novel objects, unseen object positions, different object instances or backgrounds, and longer-horizon continuous manipulation. For Bowl Stacking, the robot must additionally follow the color order specified by the language instruction.

\paragraph{Results.}
We report real-robot results under both ID and OOD settings. Across all tasks, Camera-Frame achieves the highest average success rate of 77.5\%, outperforming $\pi_{0.5}$ (70.0\%), Raw Mix (51.3\%), and From Scratch (35.0\%). This advantage is also pronounced under OOD conditions, where Camera-Frame reaches 72.5\% average success, compared with 62.5\% for $\pi_{0.5}$, 40.0\% for Raw Mix, and 22.5\% for From Scratch. These results show that camera-frame action pretraining provides a more transferable action prior and improves real-world generalization compared with naive cross-dataset action mixing.
The gains are especially clear in spatially grounded and language-conditioned tasks. For Bowl Stacking, Camera-Frame achieves 100\% success in ID and 90\% in OOD, substantially outperforming Raw Mix. Although $\pi_{0.5}$ performs better on Towel Folding, Camera-Frame still consistently surpasses From Scratch and Raw Mix. Overall, these results show that Camera-Frame provides a robust and generalizable pretraining strategy for real-world manipulation.
\section{Conclusion}
\label{sec:conclusion}
We present \system, a training-free and robot-agnostic pipeline for estimating camera extrinsics from offline robot datasets and converting heterogeneous actions into standardized camera-frame TCP actions. By providing consistent geometric semantics across different robot platforms, \system enables scalable cross-embodiment pretraining. We apply \system to 16 datasets across 4 robot platforms and show that camera-frame action pretraining improves downstream performance in both simulation and real-world experiments.

\noindent\textbf{Limitation.} Current large-scale annotation still requires occasional human intervention when tracking or mask prediction fails. We leave fully automatic offline calibration as future work.
\beginappendix
\appendix
\section{Training Details}
\label{sec:training_details}
We use StarVLA~\cite{community2026starvla} as the policy training framework for all variants. Cross-embodiment pretraining is conducted for 200K steps on 32 A100 GPUs. For both simulation and real-robot experiments, we fine-tune all variants on the target benchmark for 100K SFT steps on 8 A100 GPUs, using the same training protocol, batch size, and optimization hyperparameters for a fair comparison.

For the \textbf{Action Head baseline}, we use dataset-specific action heads in both the action encoder and decoder. Specifically, each dataset has an independent action head for mapping its original action representation into the latent action space, as well as a corresponding decoder head for mapping the latent representation back to the dataset-specific action space.

For the \textbf{Prompt baseline}, we specify the robot embodiment information in the language instruction while keeping the original action representations unchanged. Specifically, we use the following prompt template:
\begin{quote}
\small
\texttt{The robot is [robot name] with [single/dual] arm. Please predict the actions to execute the following task: [task].}
\end{quote}

\section{Dataset Overview}
\label{sec:dataset_overview}
\begin{table}[h]
\centering
\caption{Summary of annotated datasets.}
\label{tab:dataset_summary}
\resizebox{\linewidth}{!}{%
\begin{tabular}{c l l l c r}
\toprule
\# & Source & Dataset & Robot Platform & \# Third-Person Cameras & \# Samples \\
\midrule
1  & RoboMIND & ur\_1rgb & UR5e + Robotiq & 1 & 25{,}002 \\
2  & RoboMIND & franka\_3rgb & Franka & 3 & 50{,}637 \\
3  & RoboMIND & agilex\_3rgb & AgileX ALOHA & 1 & 9{,}298 \\
4  & RDT      & rdt\_aloha & AgileX ALOHA & 1 & 6{,}061 \\
5  & OXE      & nyu\_franka\_play & Franka & 2 & 730 \\
6  & OXE      & berkeley\_autolab\_ur5e & UR5e + Robotiq & 1 & 1{,}000 \\
7  & OXE      & mutex & Franka & 1 & 1{,}500 \\
8  & OXE      & toto & Franka & 1 & 901 \\
9  & OXE      & qut\_dexterous & Franka & 1 & 812 \\
10 & OXE      & franka-play & Franka & 1 & 456 \\
11 & OXE      & mimic\_play & Franka & 1 & 378 \\
12 & OXE      & sailor & Franka & 1 & 250 \\
13 & OXE      & kaist\_nonprehensile & Franka & 1 & 201 \\
14 & OXE      & ucsd\_kitchen & XArm7 & 1 & 150 \\
15 & OXE      & utokyo\_xarm & XArm7 & 1 & 92 \\
16 & OXE      & BUDS & Franka & 1 & 50 \\
\midrule
\multicolumn{5}{r}{Total} & 97{,}518 \\
\bottomrule
\end{tabular}%
}
\end{table}
Table~\ref{tab:dataset_summary} summarizes the datasets annotated by \system. 
The annotated data are collected from three major sources, including RoboMIND, RDT, and Open X-Embodiment (OXE), and cover 16 datasets across four robot platforms: Franka, UR5e, XArm7, and AgileX ALOHA. 
For each dataset, we report the data source, dataset name, robot platform, number of available third-person cameras, and number of annotated samples. 
In total, our annotation covers 97{,}518 samples, providing camera extrinsics and auxiliary annotations for both single-arm and bimanual robot datasets.

\section{Visualization of the \system Pipeline}
\begin{figure*}[t]
\centering
\includegraphics[width=\textwidth]{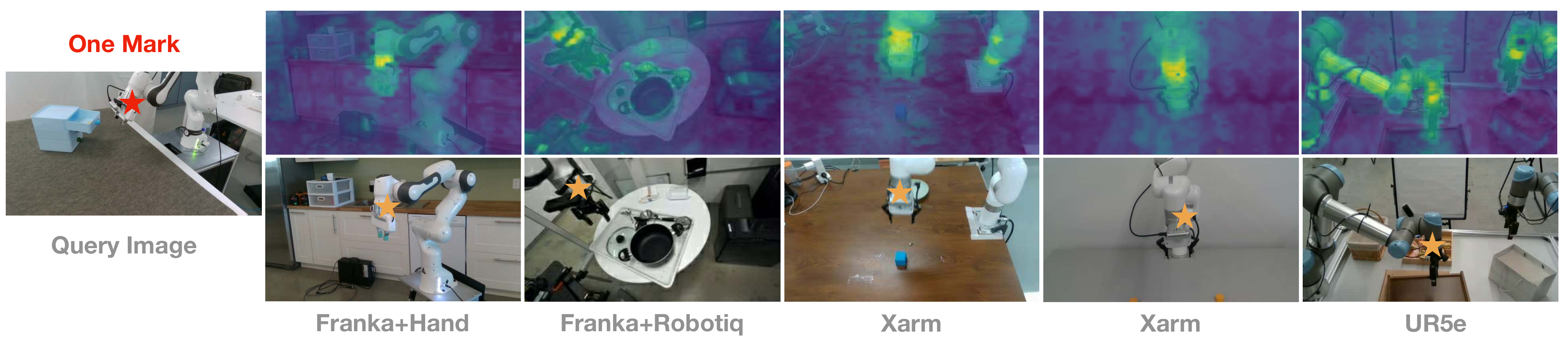}
\caption{\textbf{Qualitative result of EEF recognition on Franka, xArm and UR5e.}
The first row shows the heatmaps obtained from feature matching.
The second row visualizes the selected tracking point based on the maximum similarity.
}
\label{fig:point_result_vis}
\end{figure*}

\begin{figure*}[t]
\centering
\includegraphics[width=\textwidth]{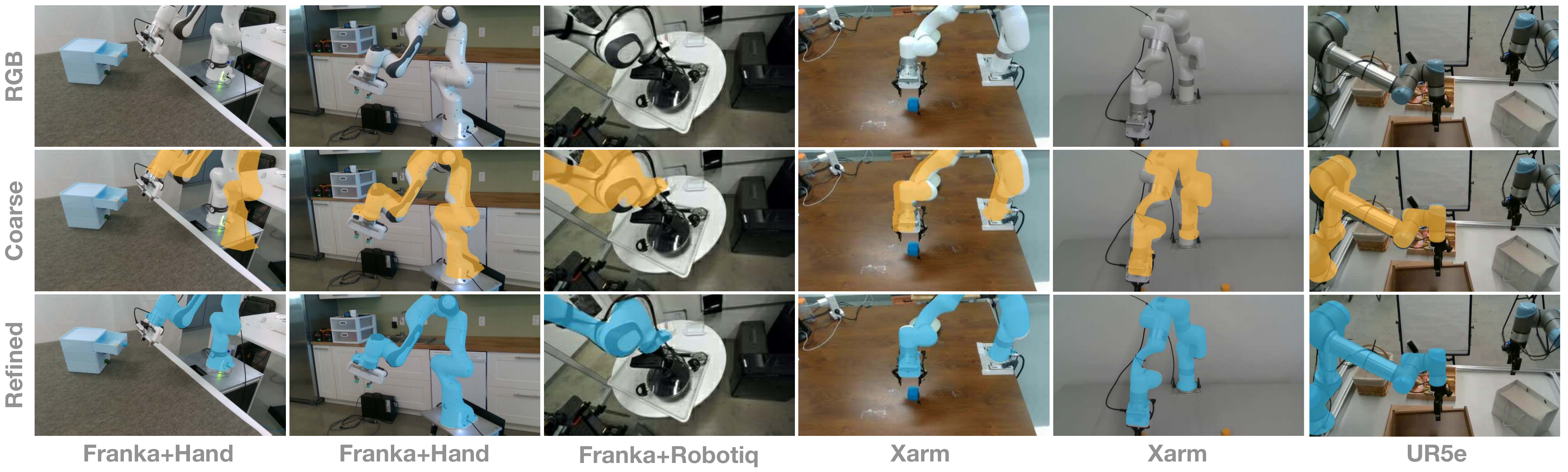}
\caption{\textbf{Qualitative result of coarse initialization and extrinsic refinement on Franka, xArm and UR5e.} The first row presents the source RGB images. The second row shows the rendered result using the camera extrinsic obtained from the automatic coarse initialization approach. The last row shows the rendered result of final camera extrinsic produced by \system.
}
\label{fig:coarse_refine}
\end{figure*}

We visualize the key stages of \system from three aspects.
First, we show the EEF recognition results across different datasets and robot platforms in~\cref{fig:point_result_vis}. 
The VFM similarity heatmaps and the selected tracking points show that a single annotated reference mark can reliably localize EEF tracking points across diverse robots.

Next, we visualize the extrinsics estimated by the coarse stage on offline datasets with different robot platforms, as shown in~\cref{fig:coarse_refine}. 
Although the rendered masks still exhibit small deviations from the observed robot masks, they already capture the camera viewpoint and robot location reasonably well, providing an effective initialization for refinement.

Finally, as shown qualitatively in~\cref{fig:coarse_refine}, the refined extrinsics produce accurate robot-image alignment across different datasets and robot platforms. 
Together, these results validate the design of \system: the coarse stage provides a stable initialization, while the refinement stage improves the estimate to achieve accurate offline camera extrinsic calibration.

\section{Calibration Experiment on DROID Dataset}
\label{sec:droid_result}
\begin{figure*}[t]
\centering
\includegraphics[width=\textwidth]{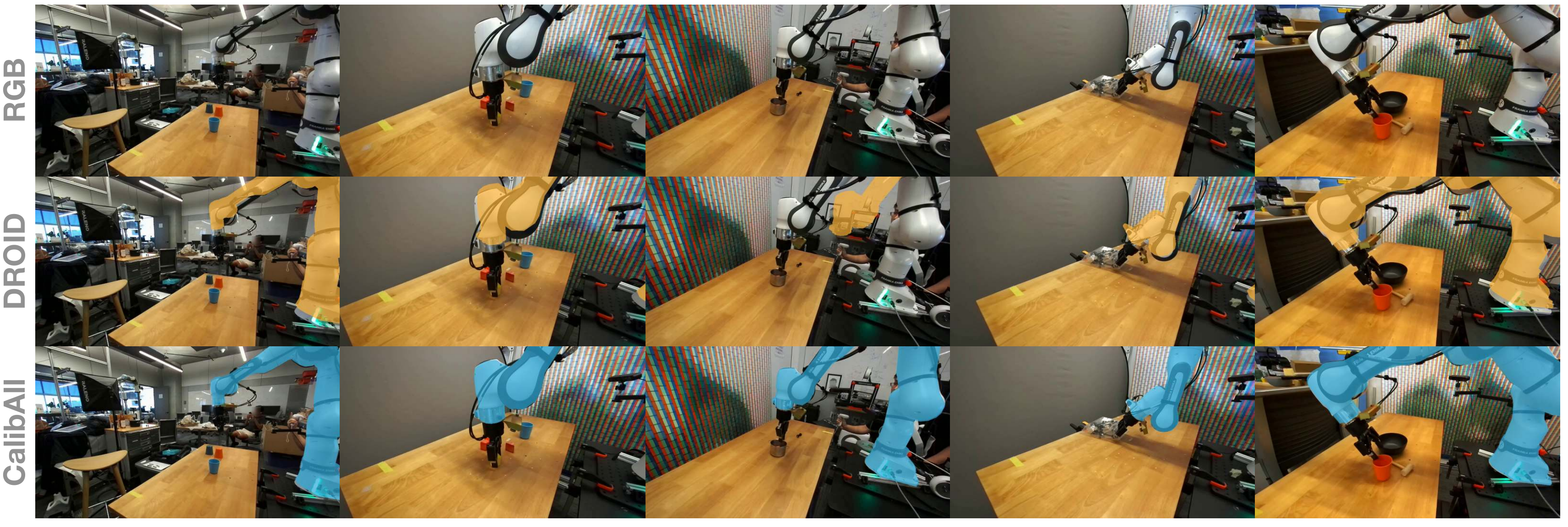}
\caption{
\textbf{Qualitative results on DROID~\cite{khazatsky2024droid}.} Rendering result using camera extrinsics in DROID~\cite{khazatsky2024droid} obtained via traditional marker-based calibration during data collection and those estimated training-free and offline by \system(ours).
}
\label{fig:comapre_droid}
\end{figure*}

\begin{table}[t]
\centering
\small
\renewcommand{\arraystretch}{1}
\caption{\textbf{Results on DROID~\cite{khazatsky2024droid}.}
We evaluate \system\ on 20 sequences.
Point denotes the success rate of end-effector mark recognition,
Track denotes the success rate of obtaining valid 2D trajectories,
and Refine denotes the success rate of camera extrinsic refinement.
$E_r$ and $E_p$ report the mean rotation and translation errors over the successful trials.}
\label{tab:droid_result}
\begin{tabular}{ccc cc}
\toprule
\multicolumn{3}{c}{Success Rate} 
& \multicolumn{2}{c}{Refinement Error} \\
\cmidrule(lr){1-3} \cmidrule(lr){4-5} 
Point & Track & Refine 
& $E_r$ (°)& $E_p$ (cm)\\
\midrule
20/20 & 15/20 & 14/20 
& 3.9 & 4.4 \\
\bottomrule
\end{tabular}
\end{table}
We conducted experiments on the more complex DROID~\cite{khazatsky2024droid} dataset. DROID is a diverse robot manipulation dataset containing 76k demonstration trajectories, or 350 hours of interaction data, collected across 564 scenes and 86 tasks. In addition, DROID records camera extrinsics for each scene obtained via traditional marker-based calibration~\cite{tsai1989new} during data collection, which solves the equation $AX=XB$.

From this dataset, we selected 20 scenes for testing. We report the success rates for each stage across these 20 scenes, as shown in the~\cref{tab:droid_result}. EEF Recognition is highly robust across all 20 scenarios, consistently achieving good initial tracking points. However, due to the complexity of the scenes and the movements of robot, tracking failed in 5 out of the 20 scenarios, causing \system\ to fail. This observation indicates that tracking performance is the main bottleneck of the current pipeline.
Nevertheless, for the 15 successful tracking scenarios, 14 achieved accurate camera extrinsic predictions. We further evaluated these 14 scenarios and found that the average position error was only 4.4 cm, with a rotation error of 3.9°. These results demonstrate the stability and effectiveness of \system\ under complex conditions.

We further visualized the masks rendered using the ground truth camera extrinsics recorded in the DROID dataset and the masks rendered using the camera extrinsics estimated by \system in~\cref{fig:comapre_droid}. The ground truth in DROID sometimes exhibits errors, whereas the results estimated by \system\ are actually more accurate and convenient than those from the DROID dataset collected using the traditional method.

\section{Real-World Experiment Results}
\begin{table}[th]
\centering
\caption{Real-world evaluation results on four tasks. We report success rates (\%) under both in-distribution (ID) and out-of-distribution (OOD) settings.}
\label{tab:real_robot_results}
\begin{tabular}{l l c c c c}
\toprule
Task & Setting & Pi-0.5 & From Scratch & Raw Mix & Camera Frame (ours) \\
\midrule
\multirow{2}{*}{Table Cleaning}
& ID  & 90 & 60 & 90 & \textbf{100} \\
& OOD & 90 & 30 & 70 & \textbf{90} \\
\midrule
\multirow{2}{*}{Pen Unscrewing}
& ID  & 60 & 50 & 60 & \textbf{70} \\
& OOD & 50 & 20 & 20 & \textbf{60} \\
\midrule
\multirow{2}{*}{Towel Folding}
& ID  & \textbf{80} & 30 & 20 & 60 \\
& OOD & \textbf{80} & 0 & 20 & 50 \\
\midrule
\multirow{2}{*}{Bowl Stacking}
& ID  & 80 & 50 & 80 & \textbf{100} \\
& OOD & 30 & 40 & 50 & \textbf{90} \\
\midrule
\multirow{2}{*}{Average}
& ID  & 77.5 & 47.5 & 62.5 & \textbf{82.5} \\
& OOD & 62.5 & 22.5 & 40.0 & \textbf{72.5} \\
\midrule
Overall Average
& -- & 70.0 & 35.0 & 51.3 & \textbf{77.5} \\
\bottomrule
\end{tabular}
\end{table}
\cref{tab:real_robot_results} reports the detailed real-world evaluation results on four tabletop tasks: Table Cleaning, Pen Unscrewing, Towel Folding, and Bowl Stacking. For each task, we evaluate both in-distribution (ID) and out-of-distribution (OOD) settings, and report the success rate over 10 trials. The OOD settings introduce variations such as novel objects, unseen object positions, different object instances or backgrounds, and longer-horizon continuous manipulation.

Camera Frame achieves the best overall average success rate of 77.5\%, outperforming Pi-0.5, From Scratch, and Raw Mix. The advantage is especially clear under OOD settings, where Camera Frame reaches 72.5\% average success, compared with 62.5\% for Pi-0.5, 40.0\% for Raw Mix, and 22.5\% for From Scratch. These results show that camera-frame action pretraining improves real-world generalization and provides a more transferable action representation across tasks.

\section{Visualization of Camera-Frame Action Dataset}
\label{sec:vis_dataset}
We visualize representative annotations produced by \system. Given the estimated camera extrinsic and robot URDF, \system renders robot masks and bounding boxes in the image space, and projects the 3D TCP trajectory into the camera view to obtain the corresponding 2D trajectory. These visualizations show that our annotations are well aligned with the observed robot across diverse platforms, camera viewpoints, and manipulation scenes, demonstrating the scalability of the Camera-Frame Action Dataset.
\begin{figure}[H]
\centering
\includegraphics[width=\linewidth]{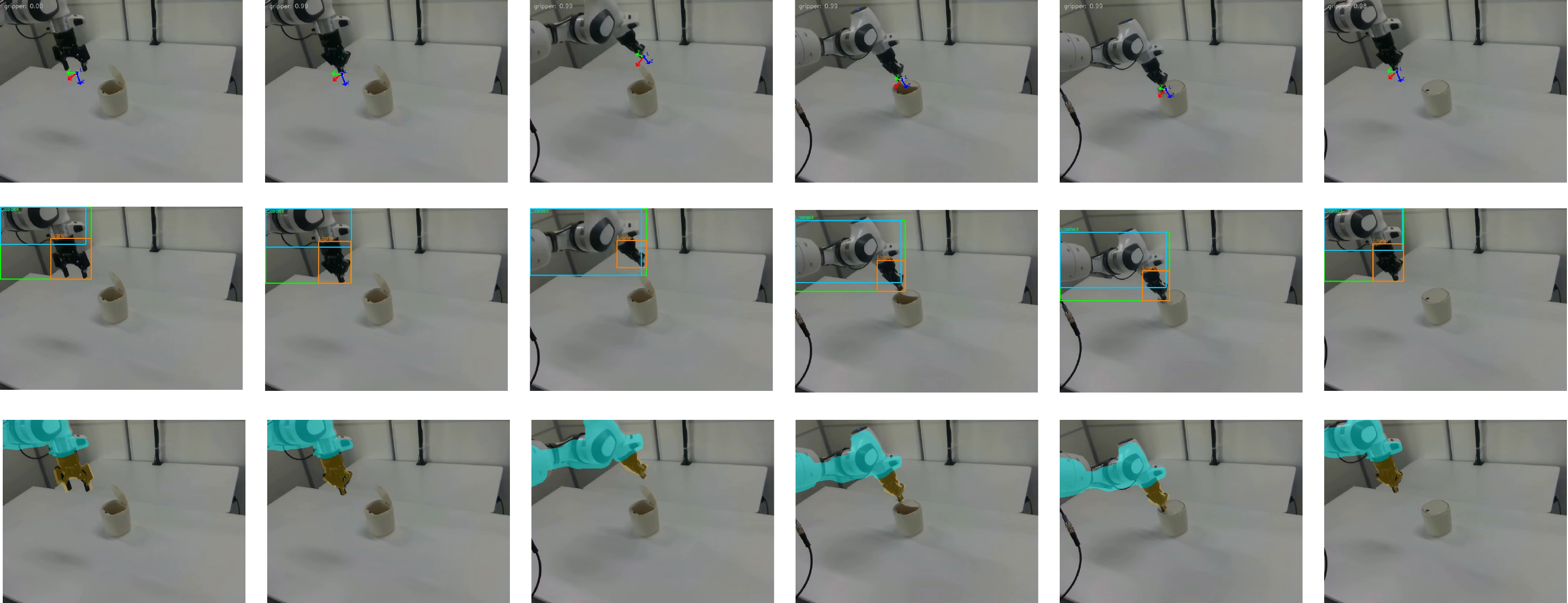}
\caption{Data Annotation on \textbf{RoboMIND Franka}.
We visualize annotations generated by \system: the first row shows 2D/3D end-effector trajectories with local coordinate axes, where red, green, and blue denote the local \textcolor{red}{$x$}, \textcolor{ForestGreen}{$y$}, and \textcolor{blue}{$z$} axes, respectively; the second row shows gripper and robot-arm bounding boxes; and the third row shows the corresponding segmentation masks.}
\end{figure}

\begin{figure}[H]
\centering
\includegraphics[width=\linewidth]{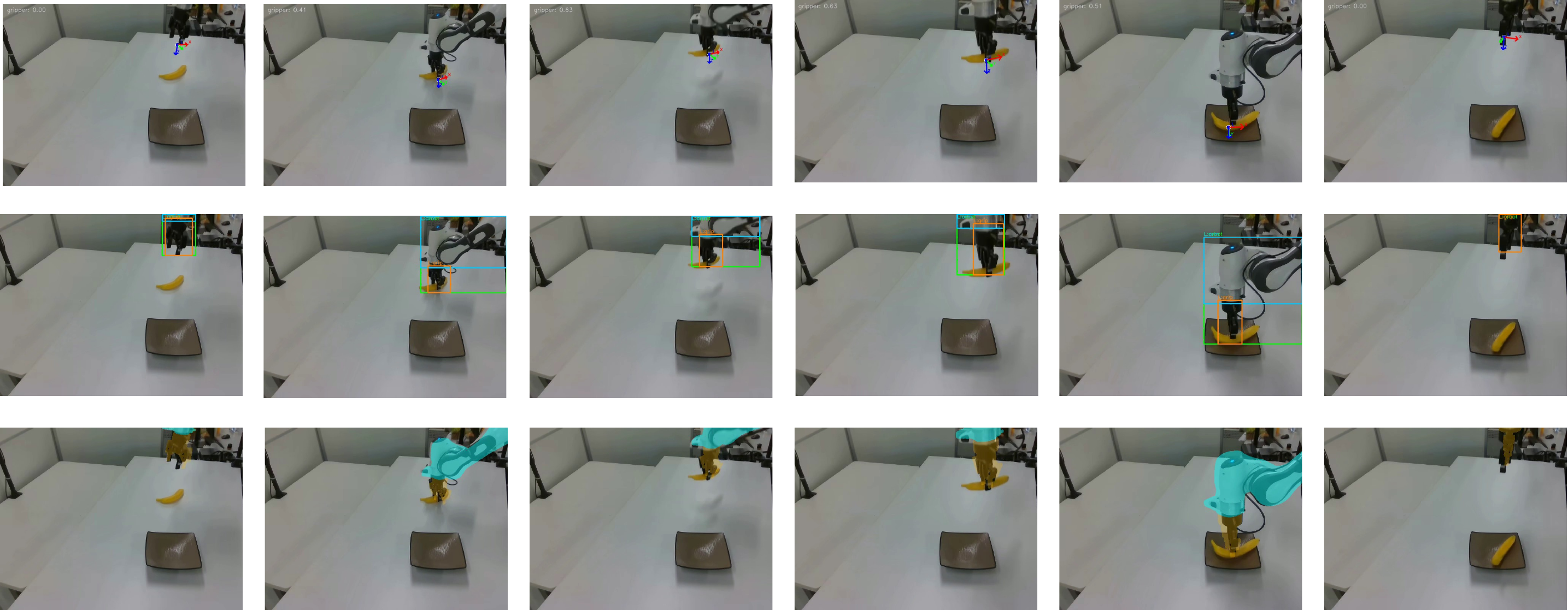}
\caption{Data Annotation on \textbf{RoboMIND Franka}.
We visualize annotations generated by \system: the first row shows 2D/3D end-effector trajectories with local coordinate axes, where red, green, and blue denote the local \textcolor{red}{$x$}, \textcolor{ForestGreen}{$y$}, and \textcolor{blue}{$z$} axes, respectively; the second row shows gripper and robot-arm bounding boxes; and the third row shows the corresponding segmentation masks.}
\end{figure}

\begin{figure}[H]
\centering
\includegraphics[width=\linewidth]{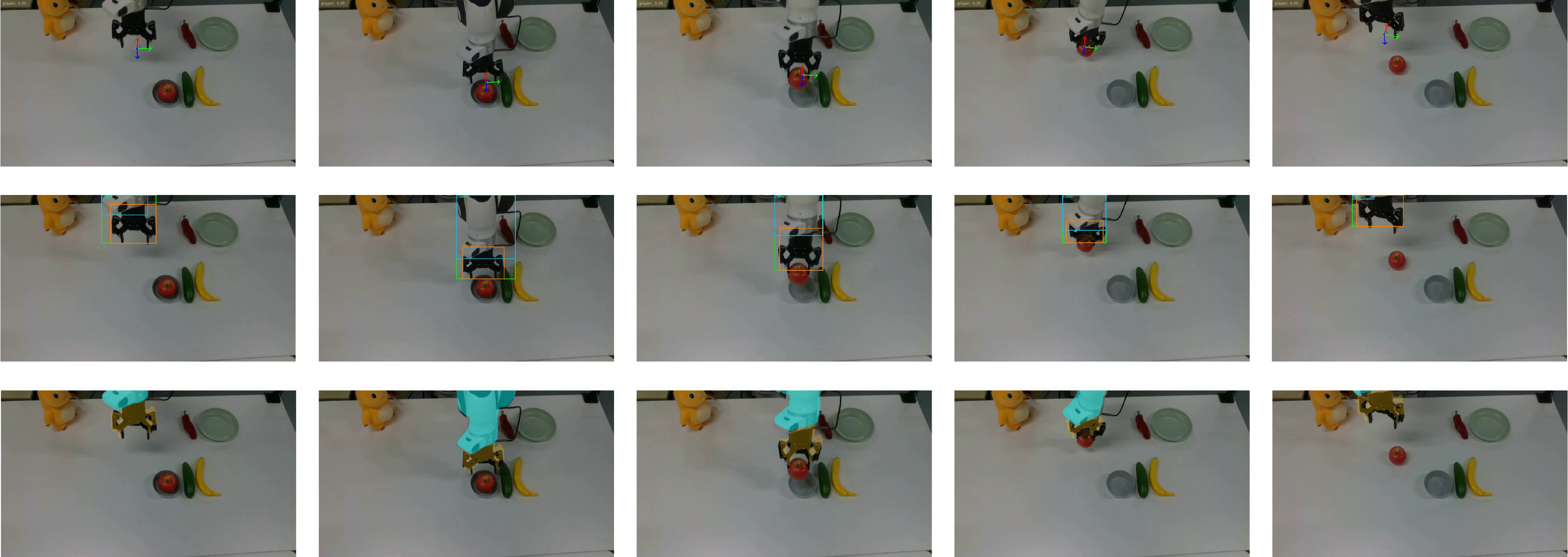}
\caption{Data Annotation on \textbf{RoboMIND Franka}.
We visualize annotations generated by \system: the first row shows 2D/3D end-effector trajectories with local coordinate axes, where red, green, and blue denote the local \textcolor{red}{$x$}, \textcolor{ForestGreen}{$y$}, and \textcolor{blue}{$z$} axes, respectively; the second row shows gripper and robot-arm bounding boxes; and the third row shows the corresponding segmentation masks.}
\end{figure}

\begin{figure}[H]
\centering
\includegraphics[width=\linewidth]{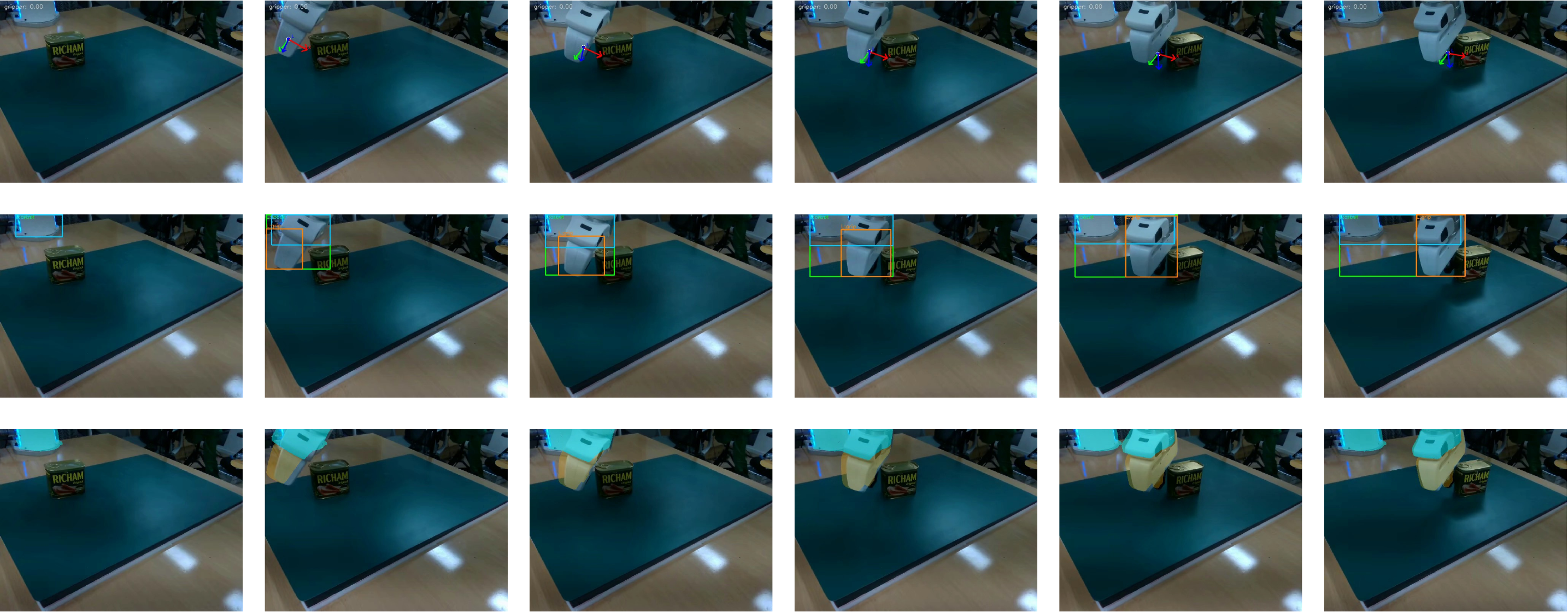}
\caption{Data Annotation on \textbf{KAIST}.
We visualize annotations generated by \system: the first row shows 2D/3D end-effector trajectories with local coordinate axes, where red, green, and blue denote the local \textcolor{red}{$x$}, \textcolor{ForestGreen}{$y$}, and \textcolor{blue}{$z$} axes, respectively; the second row shows gripper and robot-arm bounding boxes; and the third row shows the corresponding segmentation masks.}
\end{figure}

\begin{figure}[H]
\centering
\includegraphics[width=\linewidth]{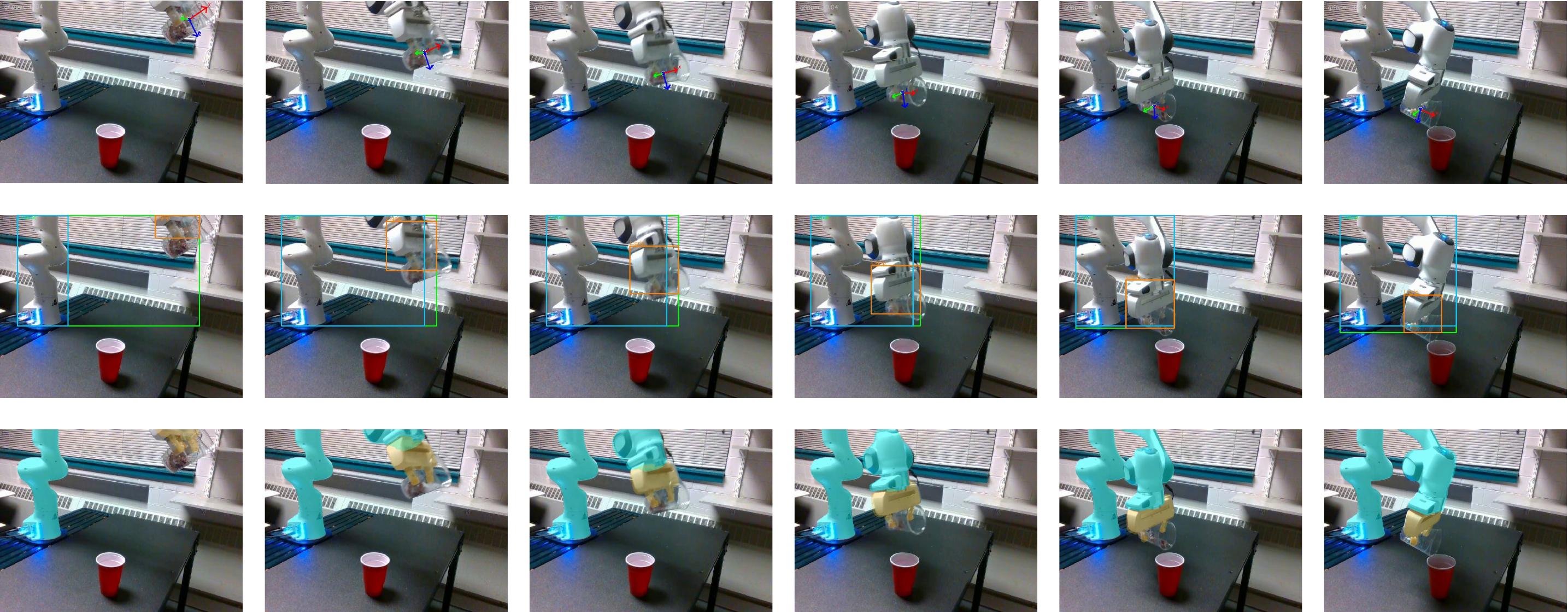}
\caption{Data Annotation on \textbf{TOTO}.
We visualize annotations generated by \system: the first row shows 2D/3D end-effector trajectories with local coordinate axes, where red, green, and blue denote the local \textcolor{red}{$x$}, \textcolor{ForestGreen}{$y$}, and \textcolor{blue}{$z$} axes, respectively; the second row shows gripper and robot-arm bounding boxes; and the third row shows the corresponding segmentation masks.}
\end{figure}

\begin{figure}[H]
\centering
\includegraphics[width=\linewidth]{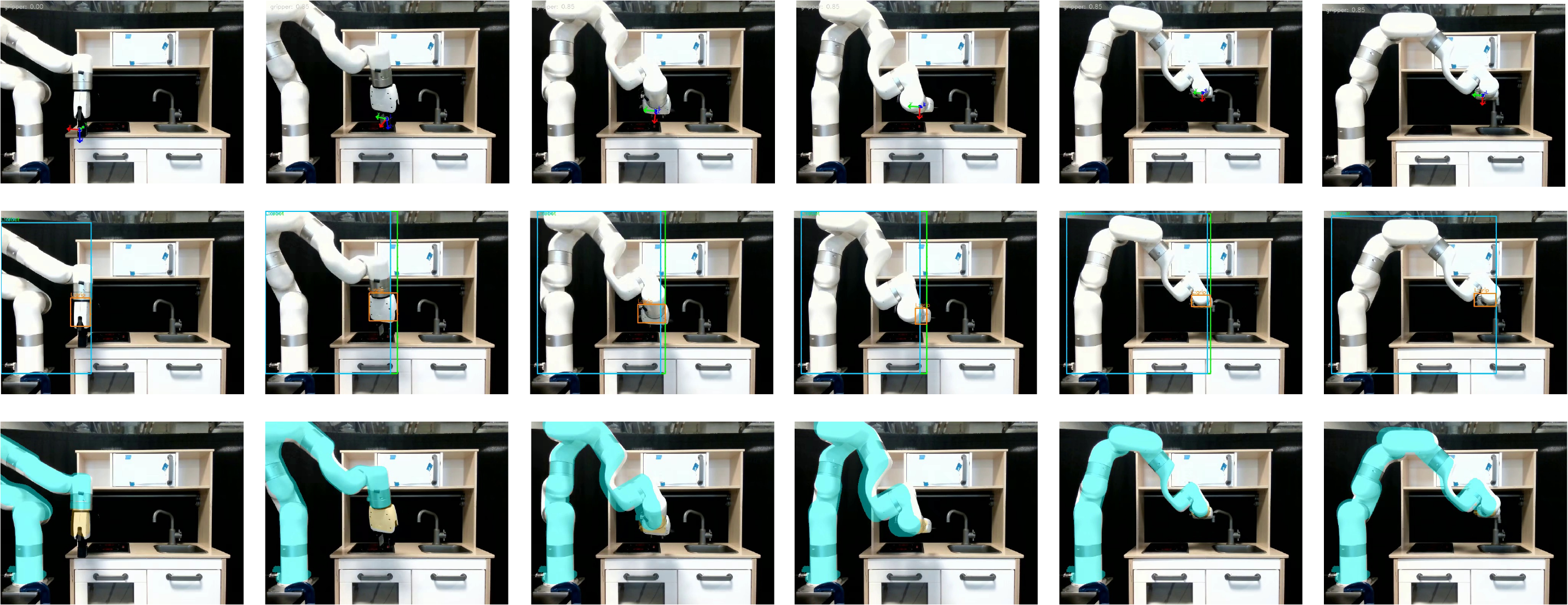}
\caption{Data Annotation on \textbf{UCSD Kitchen}.
We visualize annotations generated by \system: the first row shows 2D/3D end-effector trajectories with local coordinate axes, where red, green, and blue denote the local \textcolor{red}{$x$}, \textcolor{ForestGreen}{$y$}, and \textcolor{blue}{$z$} axes, respectively; the second row shows gripper and robot-arm bounding boxes; and the third row shows the corresponding segmentation masks.}
\end{figure}

\begin{figure}[H]
\centering
\includegraphics[width=\linewidth]{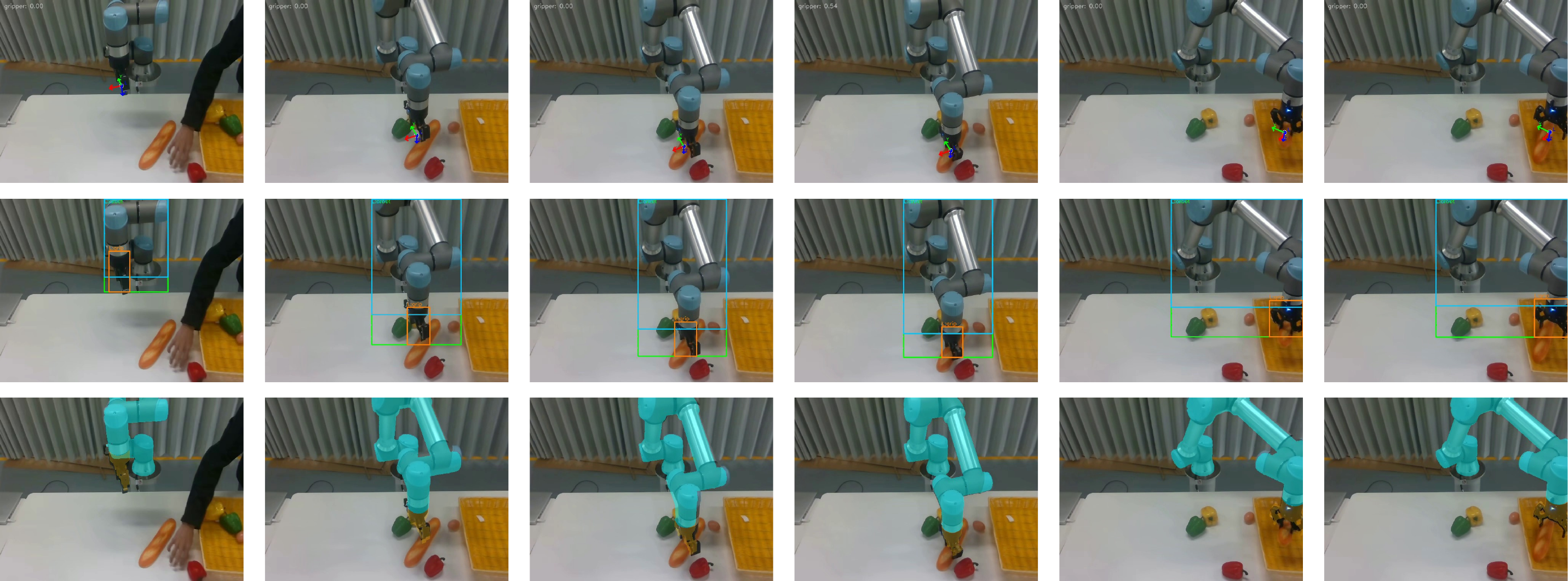}
\caption{Data Annotation on \textbf{RoboMIND UR5e}.
We visualize annotations generated by \system: the first row shows 2D/3D end-effector trajectories with local coordinate axes, where red, green, and blue denote the local \textcolor{red}{$x$}, \textcolor{ForestGreen}{$y$}, and \textcolor{blue}{$z$} axes, respectively; the second row shows gripper and robot-arm bounding boxes; and the third row shows the corresponding segmentation masks.}
\end{figure}

\begin{figure}[H]
\centering
\includegraphics[width=\linewidth]{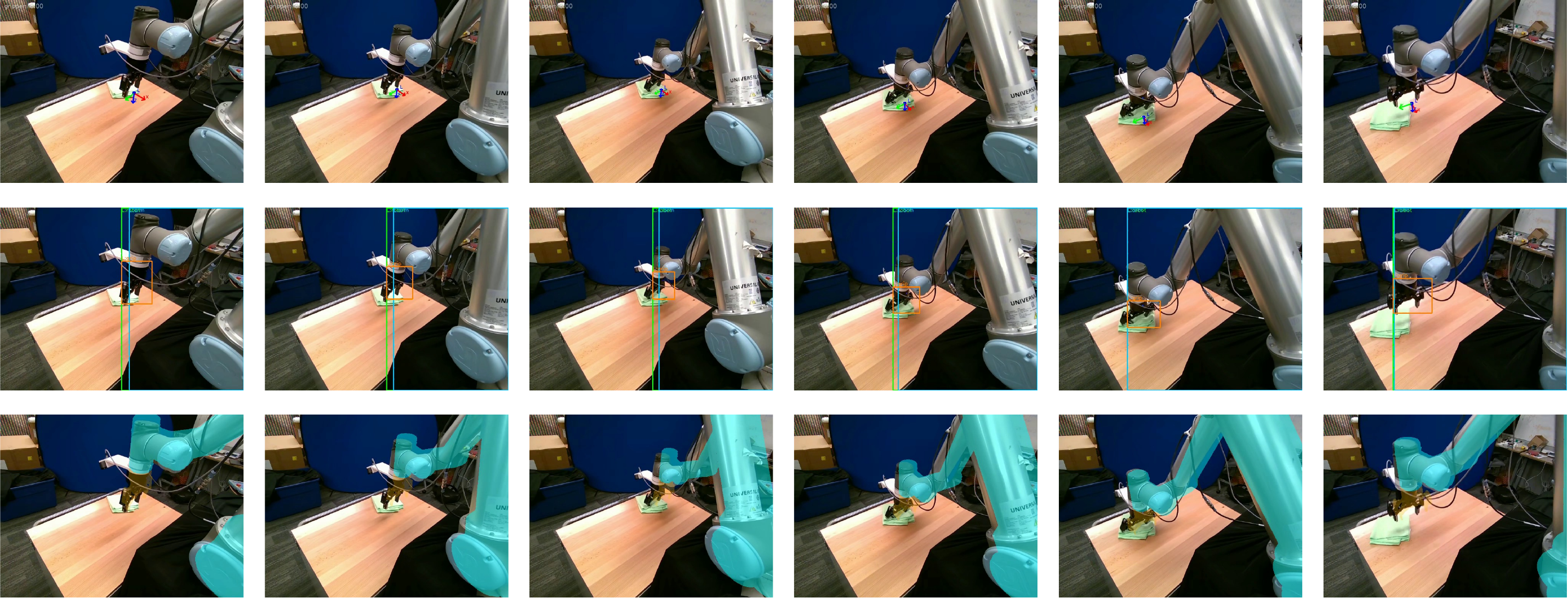}
\caption{Data Annotation on \textbf{Berkeley UR5}.
We visualize annotations generated by \system: the first row shows 2D/3D end-effector trajectories with local coordinate axes, where red, green, and blue denote the local \textcolor{red}{$x$}, \textcolor{ForestGreen}{$y$}, and \textcolor{blue}{$z$} axes, respectively; the second row shows gripper and robot-arm bounding boxes; and the third row shows the corresponding segmentation masks.}
\end{figure}

\begin{figure}[H]
\centering
\includegraphics[width=\linewidth]{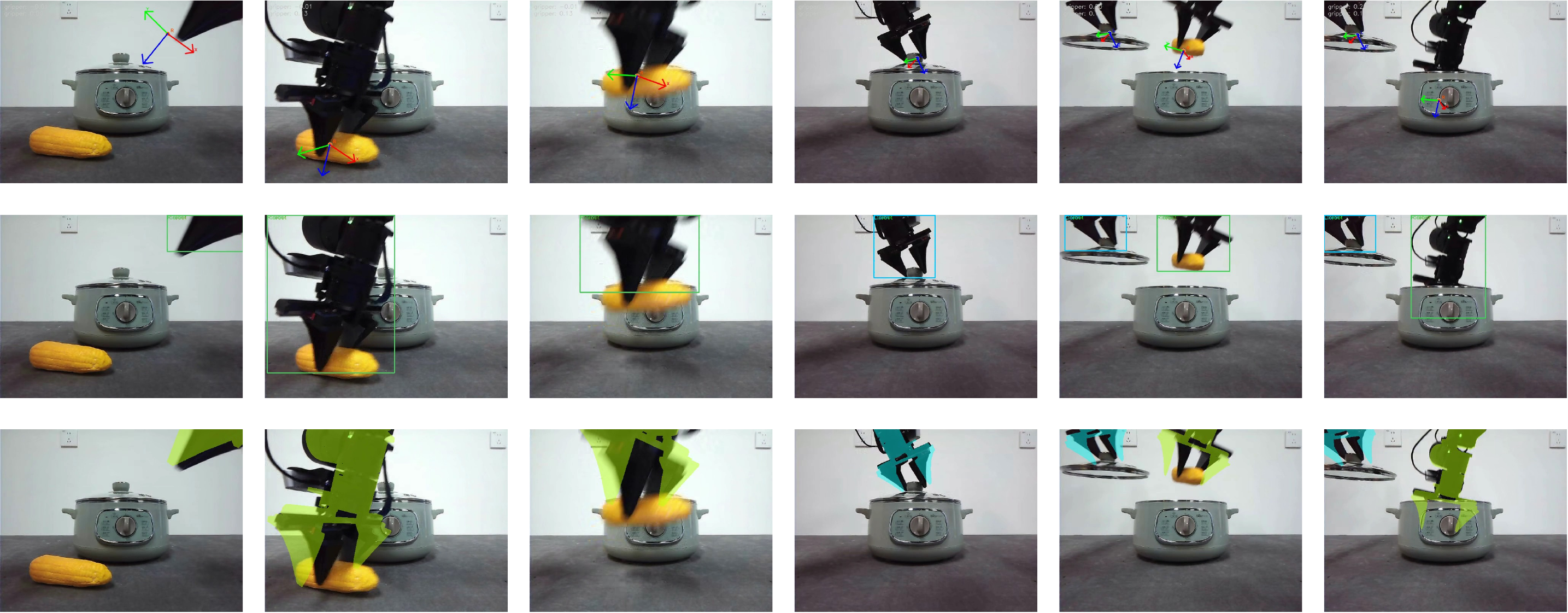}
\caption{Data Annotation on \textbf{RoboMIND ALOHA}.
We visualize annotations generated by \system: the first row shows 2D/3D end-effector trajectories with local coordinate axes, where red, green, and blue denote the local \textcolor{red}{$x$}, \textcolor{ForestGreen}{$y$}, and \textcolor{blue}{$z$} axes, respectively; the second row shows gripper and robot-arm bounding boxes; and the third row shows the corresponding segmentation masks.}
\end{figure}

\begin{figure}[H]
\centering
\includegraphics[width=\linewidth]{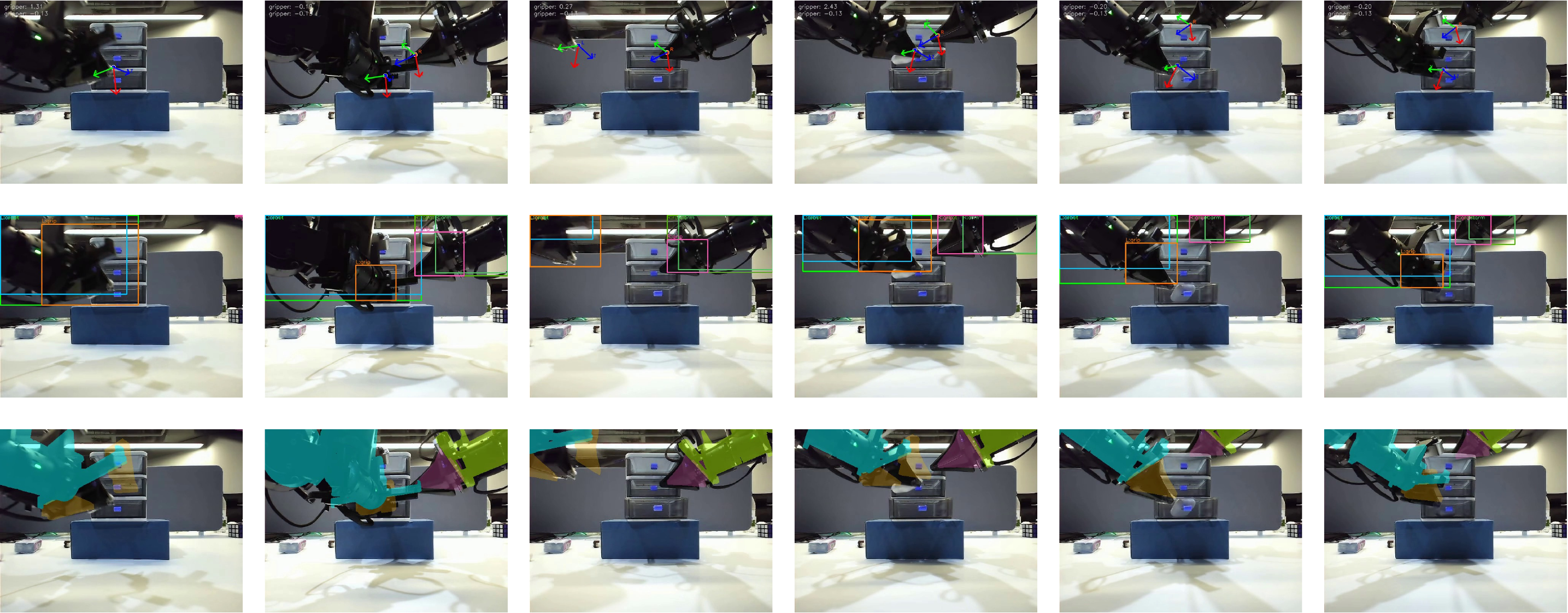}
\caption{Data Annotation on \textbf{RDT}.
We visualize annotations generated by \system: the first row shows 2D/3D end-effector trajectories with local coordinate axes, where red, green, and blue denote the local \textcolor{red}{$x$}, \textcolor{ForestGreen}{$y$}, and \textcolor{blue}{$z$} axes, respectively; the second row shows gripper and robot-arm bounding boxes; and the third row shows the corresponding segmentation masks.}
\end{figure}

\clearpage

\bibliographystyle{plainnat}
\bibliography{main}

\clearpage

\end{document}